\documentclass[10pt,twocolumn,letterpaper]{article}
\usepackage{iccv}
\usepackage{times}
\usepackage{graphicx}
\usepackage{amsmath}
\usepackage{amssymb}
\usepackage{lipsum}
\usepackage{picture}
\usepackage{stmaryrd}
\usepackage[%
pagebackref=true,
breaklinks=true,
letterpaper=true,
colorlinks,
bookmarks=false]{hyperref}
\usepackage{cleveref}
\usepackage{multirow}
\usepackage{diagbox}

\crefname{section}{sec.}{secs.}
\Crefname{section}{Sec.}{Secs.}

\usepackage[textwidth=1.7cm,textsize=tiny]{todonotes}
\graphicspath{{./figures/}}

\IfFileExists{/Users/vedaldi/.bashrc}{\iccvfinalcopy}{}
\iccvfinalcopy 
\def\iccvPaperID{1393} 
\ificcvfinal\fi

\title{Interpretable Explanations of Black Boxes by Meaningful Perturbation}
\author{Ruth C. Fong\\
University of Oxford\\
{\tt\small ruthfong@robots.ox.ac.uk}
\and
Andrea Vedaldi\\
University of Oxford\\
{\tt\small vedaldi@robots.ox.ac.uk}}

\begin{document}
\maketitle

\begin{abstract}
As machine learning algorithms are increasingly applied to high impact yet high risk tasks, such as medical diagnosis or autonomous driving, it is critical that researchers can explain how such algorithms arrived at their predictions. In recent years, a number of image saliency methods have been developed to summarize where highly complex neural networks ``look'' in an image for evidence for their predictions. However, these techniques are limited by their heuristic nature and architectural constraints.

In this paper, we make two main contributions: First, we propose a general framework for learning different kinds of explanations for any black box algorithm. Second, we specialise the framework to find the part of an image most responsible for a classifier decision. Unlike previous works, our method is model-agnostic and testable because it is grounded in explicit and interpretable image perturbations.
\end{abstract}
\vspace{-1em}

\section{Introduction}\label{s:intro}

Given the powerful but often opaque nature of modern black box predictors such as deep neural networks~\cite{krizhevsky2012imagenet,kurakin2016adversarial}, there is a considerable interest in \emph{explaining} and \emph{understanding} predictors \emph{a-posteriori}, after they have been learned. This remains largely an open problem. One reason is that we lack a formal understanding of what it means to explain a classifier. Most of the existing approaches~\cite{zeiler2014visualizing,springenberg2014striving,mahendran2016salient,mahendran2015understanding,mahendran2016visualizing,zeiler2014visualizing}, etc., often produce intuitive visualizations; however, since such visualizations are primarily heuristic, their meaning remains unclear.

In this paper, we revisit the concept of ``explanation'' at a formal level, with the goal of developing principles and methods to explain  any black box function $f$, e.g. a neural network object classifier. Since such a function is learned automatically from data, we would like to understand \emph{what} $f$ has learned to do and \emph{how} it does it. Answering the ``what'' question means determining the properties of the map. The ``how'' question investigates the internal mechanisms that allow the map to achieve these properties.
We focus mainly on the ``what'' question and argue that it can be answered by providing \emph{interpretable rules} that describe the input-output relationship captured by $f$. For example, one rule could be that $f$ is rotation invariant, in the sense that ``$f(x)=f(x')$ whenever images $x$ and $x'$ are related by a rotation''.

\begin{figure}[t]
\begin{center}
  \includegraphics[width=\linewidth, trim=0 .9em 0 .9em]{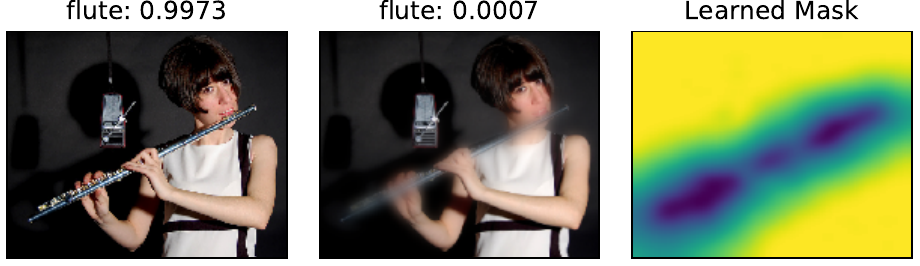}
\end{center}
\caption{An example of a mask learned (right) by blurring an image (middle) to suppress the softmax probability of its target class (left: original image; softmax scores above images).}
\label{f:splash}
\end{figure}


In this paper, we make several contributions. First, we propose the general framework of explanations as meta-predictors (\cref{s:related}), extending~\cite{turner15a-model}'s work. Second, we identify several pitfalls in designing automatic explanation systems. We show in particular that neural network artifacts are a major attractor for explanations. While artifacts are informative since they explain part of the network behavior, characterizing other properties of the network requires careful calibration of the \emph{generality} and \emph{interpretability} of explanations. Third, we reinterpret network saliency in our framework. We show that this provides a natural generalization of the gradient-based saliency technique of~\cite{simonyan14deep} by \emph{integrating} information over several rounds of backpropagation in order to learn an explanation. We also compare this technique to other methods~\cite{simonyan14deep, springenberg2014striving,zhang2016top,selvaraju2016grad,zeiler2014visualizing} in terms of their meaning and obtained results.

\section{Related work}\label{s:related}
Our work builds on~\cite{simonyan14deep}'s gradient-based method, which backpropagates the gradient for a class label to the image layer. Other 
backpropagation methods include DeConvNet~\cite{zeiler2014visualizing} and Guided Backprop~\cite{springenberg2014striving,mahendran2016salient}, which builds off of DeConvNet~\cite{zeiler2014visualizing} and~\cite{simonyan14deep}'s gradient method to produce sharper visualizations. 

Another set of techniques incorporate network activations into their visualizations: Class Activation Mapping (CAM)~\cite{zhou2016learning} and its  relaxed generalization Grad-CAM~\cite{selvaraju2016grad} visualize the linear combination of a late layer's activations and class-specific weights (or gradients for~\cite{selvaraju2016grad}), while Layer-Wise Relevance Propagation (LRP)~\cite{bach2015pixel} and Excitation Backprop~\cite{zhang2016top} backpropagate an class-specific error signal though a network while multiplying it with each convolutional layer's activations.

With the exception of~\cite{simonyan14deep}'s gradient method, the above techniques introduce different backpropagation heuristics, which results in aesthetically pleasing but heuristic notions of image saliency. They also are not model-agnostic, with most being limited to neural networks (all except~\cite{simonyan14deep,bach2015pixel}) and many requiring architectural modifications~\cite{zeiler2014visualizing,springenberg2014striving,mahendran2016salient,zhou2016learning} and/or access to intermediate layers~\cite{zhou2016learning,selvaraju2016grad,bach2015pixel,zhang2016top}.

A few techniques examine the relationship between inputs and outputs by editing an input image and observing its effect on the output. These include greedily graying out segments of an image until it is misclassified ~\cite{zhou2014object} and visualizing the classification score drop when an image is occluded at fixed regions~\cite{zeiler2014visualizing}. However, these techniques are limited by their approximate nature; we introduce a differentiable method that allows for the effect of the joint inclusion/exclusion of different image regions to be considered.

Our research also builds on the work of~\cite{turner15a-model,ribeiro2016should,cao2015look}. The idea of explanations as predictors is inspired by the work of~\cite{turner15a-model}, which we generalize to new types of explanations, from classification to invariance. 

The Local Intepretable Model-Agnostic Explanation (LIME) framework~\cite{ribeiro2016should} is relevant to our local explanation paradigm and saliency method (sections~\ref{s:local},~\ref{s:sal}) in that both use an function's output with respect to inputs from a neighborhood around an input $x_0$ that are generated by perturbing the image. However, their method takes much longer to converge ($N = 5000$ vs. our $300$ iterations) and produces a coarse heatmap defined by fixed super-pixels.


Similar to how our paradigm aims to learn an image perturbation mask that minimizes a class score, feedback networks~\cite{cao2015look} learn gating masks after every ReLU in a network to maximize a class score. However, our masks are plainly interpretable as they directly edit the image while ~\cite{cao2015look}'s ReLU gates are not and can not be directly used as a visual explanation; furthermore, their method requires architectural modification and may yield different results for different networks, while ours is model-agnostic.


\begin{figure*}\centering
\includegraphics[width=\linewidth]{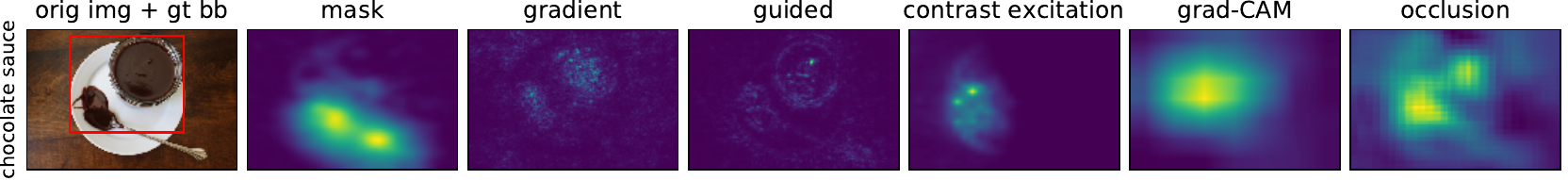}
\includegraphics[width=\linewidth]{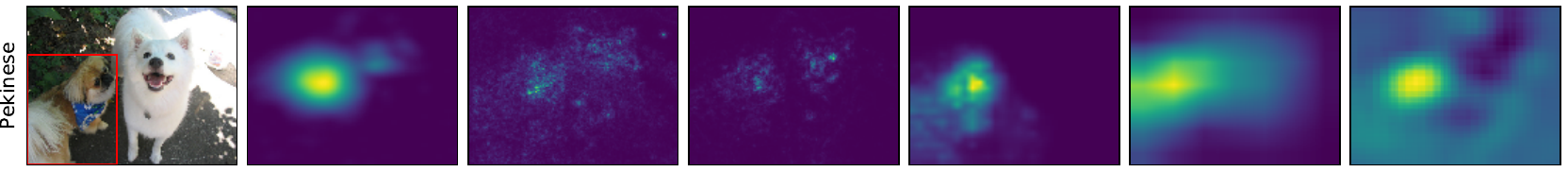}
\includegraphics[width=\linewidth]{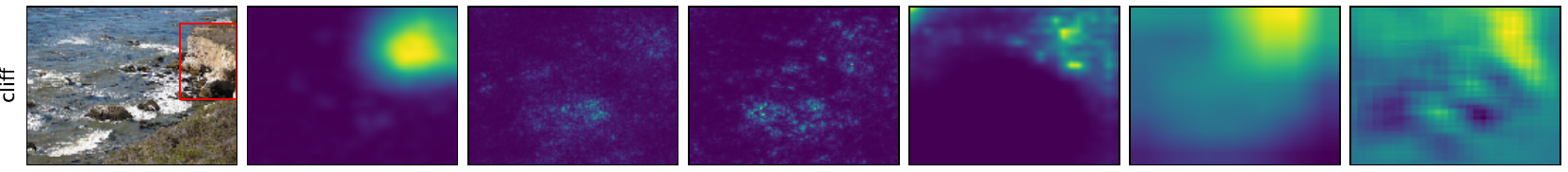}
\includegraphics[width=\linewidth]{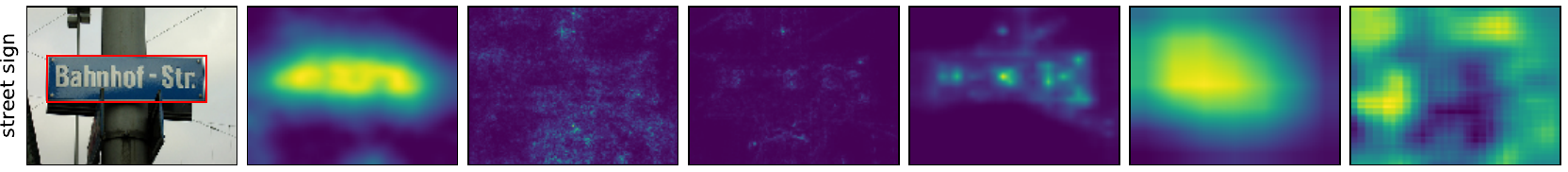}
\includegraphics[width=\linewidth]{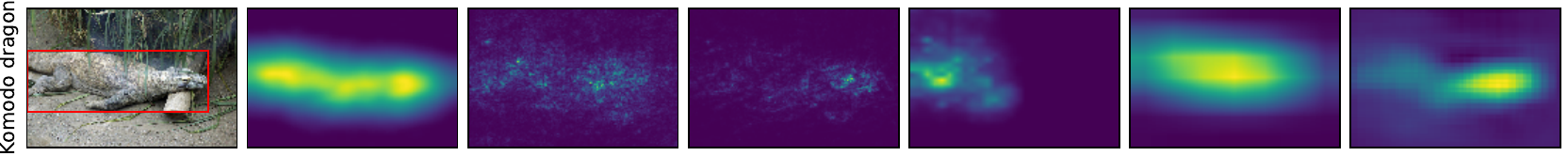}
\includegraphics[width=\linewidth]{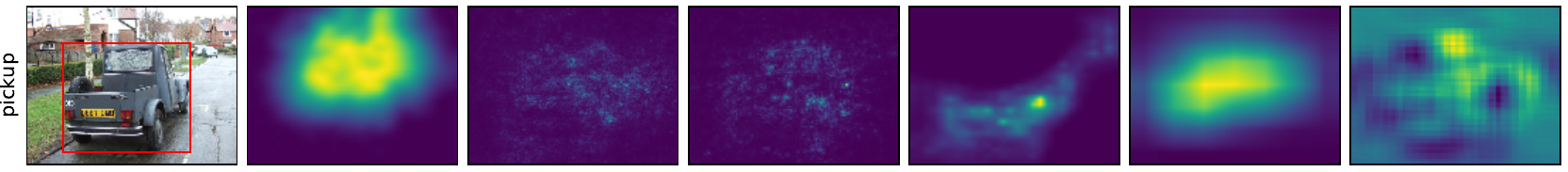}
\includegraphics[width=\linewidth]{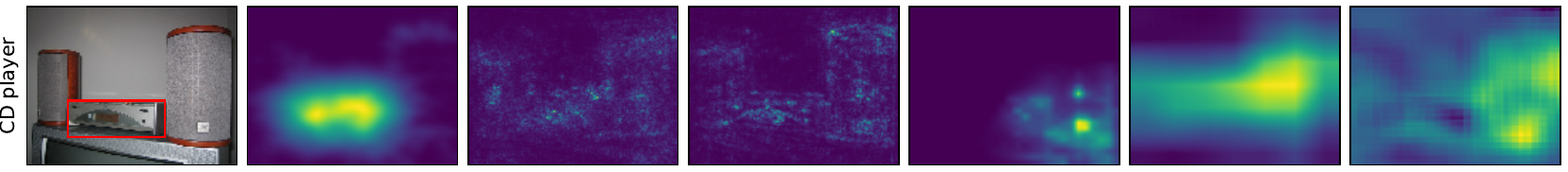}
\includegraphics[width=\linewidth]{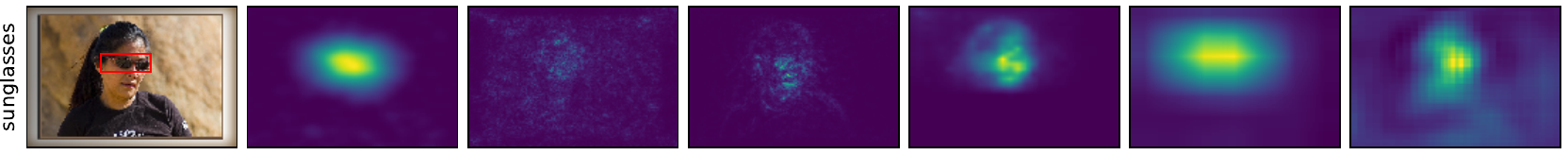}
\includegraphics[width=\linewidth]{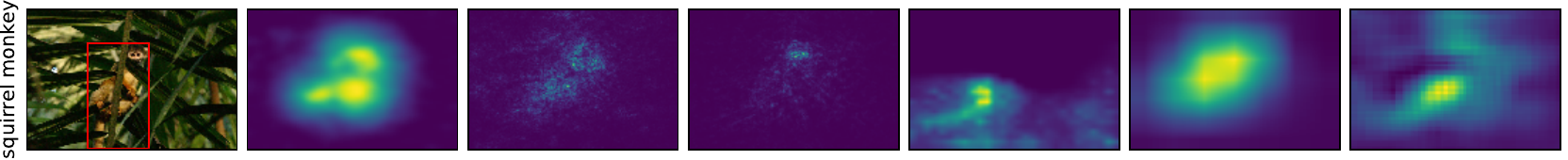}
\includegraphics[width=\linewidth]{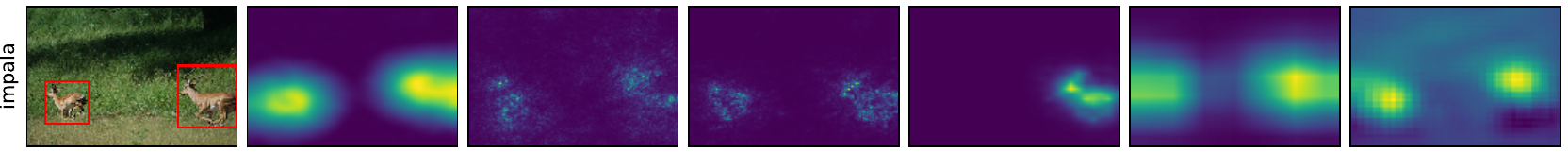}
\includegraphics[width=\linewidth]{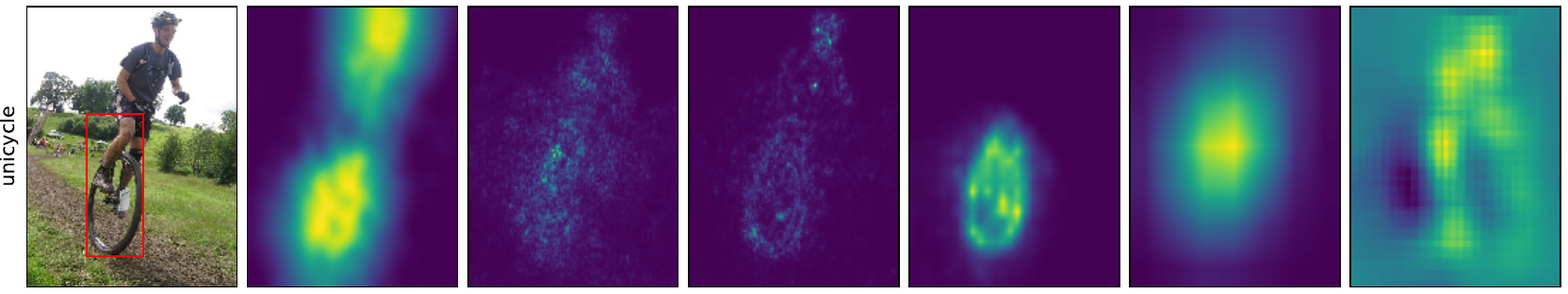}
   \caption{Comparison with other saliency methods. From left to right: original image with ground truth bounding box, learned mask subtracted from 1 (our method), gradient-based saliency~\cite{simonyan14deep}, guided backprop~\cite{springenberg2014striving,mahendran2016salient}, contrastive excitation backprop~\cite{zhang2016top}, Grad-CAM~\cite{selvaraju2016grad}, and occlusion~\cite{zeiler2014visualizing}.}
\label{f:comparison}
\end{figure*}
\section{Explaining black boxes with meta-learning}\label{s:method}

A \emph{black box} is a map $f: \mathcal{X}\rightarrow\mathcal{Y}$ from an input space $\mathcal{X}$ to an output space $\mathcal{Y}$, typically obtained from an opaque learning process. To make the discussion more concrete, consider as input color images $x : \Lambda \rightarrow \mathbb{R}^3$ where $\Lambda=\{1,\dots,H\}\times\{1,\dots,W\}$ is a discrete domain. The output $y\in\mathcal{Y}$ can be a boolean $\{-1,+1\}$ telling whether the image contains an object of a certain type (e.g.\ a \emph{robin}), the probability of such an event, or some other interpretation of the image content.


\subsection{Explanations as meta-predictors}\label{s:meta}

An \emph{explanation} is a rule that predicts the response of a black box $f$ to certain inputs. For example, we can explain a behavior of a \emph{robin} classifier by the rule $Q_1(x;f) = \{ x \in\mathcal{X}_c  \Leftrightarrow  f(x) = +1 \}$, where $\mathcal{X}_c \subset \mathcal{X}$ is the subset of all the \emph{robin} images. Since $f$ is imperfect, any such rule applies only approximately. We can measure the faithfulness of the explanation as its expected prediction error: $\mathcal{L}_1 = \mathbb{E}[1 - \delta_{Q_1(x;f)}]$, where $\delta_Q$ is the indicator function of event $Q$. Note that $Q_1$ implicitly requires a distribution $p(x)$ over possible images $\mathcal{X}$. Note also that $\mathcal{L}_1$ is simply the expected prediction error of the classifier. Unless we did not know that $f$ was trained as a \emph{robin} classifier, $Q_1$ is not very insightful, but it is interpretable since $\mathcal{X}_c$ is.

Explanations can also make relative statements about black box outcomes. For example, a black box $f$, could be rotation invariant: $Q_2(x,x';f) = \{ x \sim_\text{rot} x' \Rightarrow f(x) = f(x') \}$, where $x \sim_\text{rot} x'$ means that $x$ and $x'$ are related by a rotation. Just like before, we can measure the faithfulness of this explanation as $\mathcal{L}_2 = \mathbb{E}[1-\delta_{Q_2(x,x';f)}|x \sim x']$.\footnote{For rotation invariance we condition on $x \sim x'$ because the probability of independently sampling rotated $x$ and $x'$ is zero, so that, without conditioning, $Q_2$ would be true with probability 1.} This rule is interpretable because the relation $\sim_\text{rot}$ is.


\paragraph{Learning explanations.} A significant advantage of formulating explanations as meta predictors is that their faithfulness can be measured as prediction accuracy. Furthermore, machine learning algorithms can be used to \emph{discover explanations} automatically, by finding explanatory rules $Q$ that apply to a certain classifier $f$ out of a large pool of possible rules $\mathcal{Q}$.

In particular, finding the \emph{most accurate} explanation $Q$ is similar to a traditional learning problem and can be formulated computationally as a \emph{regularized empirical risk minimization} such as:
\begin{equation}\label{e:learn}
   \min_{Q\in\mathcal{Q}} \lambda\mathcal{R}(Q) + \frac{1}{n}\sum_{i=1}^n \mathcal{L}(Q(x_i;f),x_i,f),
   \ x_i \sim p(x).
\end{equation}
Here, the regularizer $\mathcal{R}(Q)$ has two goals: to allow the explanation $Q$ to generalize beyond the $n$ samples $x_1,\dots,x_n$ considered in the optimization and to pick an explanation $Q$ which is simple and thus, hopefully, more interpretable.

\paragraph{Maximally informative explanations.} Simplicity and interpretability are often not sufficient to find good explanations and must be paired with informativeness. Consider the following variant of rule $Q_2$: $Q_3(x,x';f,\theta) = \{ x\sim_\theta x' \Rightarrow f(x) = f(x') \}$, where $x \sim_\theta x'$ means that $x$ and $x'$ are related by a rotation of an angle $\leq \theta$. Explanations for larger angles imply the ones for smaller ones, with $\theta=0$ being trivially satisfied. The regularizer $\mathcal{R}(Q_3(\cdot;\theta))=-\theta$ can then be used to select a maximal angle and thus find an explanation that is as informative as possible.\footnote{Naively, strict invariance for any $\theta >0$ implies invariance to arbitrary rotations as small rotations compose into larger ones. However, the formulation can still be used to describe rotation insensitivity (when $f$ varies slowly with rotation), or $\sim_\theta$'s meaning can be changed to indicate rotation w.r.t.\ a canonical ``upright'' direction for a certain object classes, etc.}

\subsection{Local explanations}\label{s:local}

\begin{figure}\centering
  \includegraphics[width=0.45\linewidth]{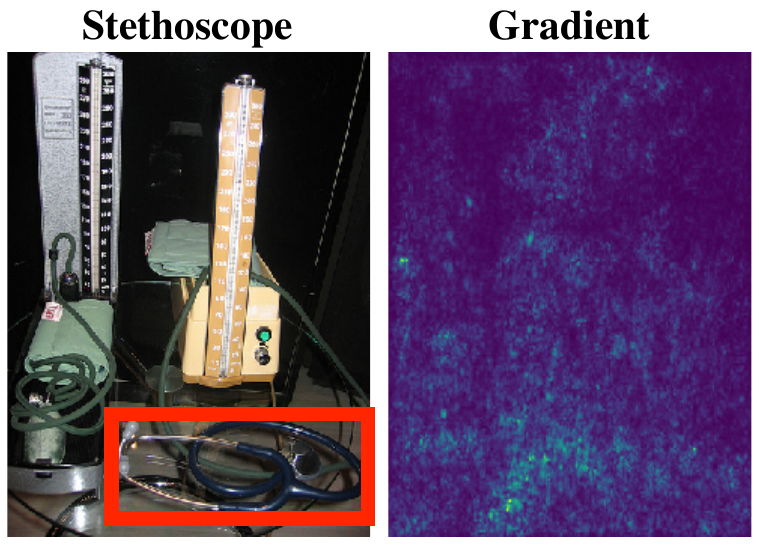}
  \includegraphics[width=0.45\linewidth]{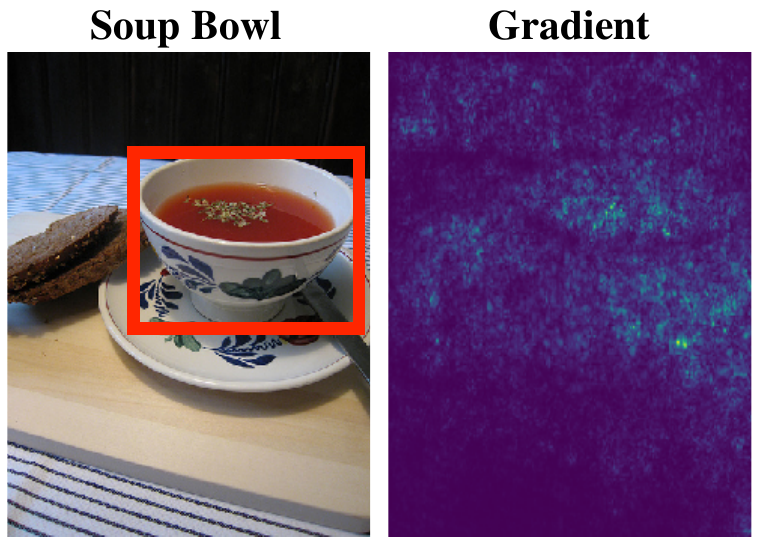}
\caption{Gradient saliency maps of~\cite{simonyan14deep}. A red bounding box highlight the object which is meant to be recognized in the image. Note the strong response in apparently non-relevant image regions.}
\label{f:simonyan}
\end{figure}

A \emph{local explanation} is a rule $Q(x;f,x_0)$ that predicts the response of $f$ in a neighborhood of a certain point $x_0$. If $f$ is smooth at $x_0$, it is natural to construct $Q$ by using the first-order Taylor expansion of $f$:
\begin{equation}\label{e:simo}
f(x)\approx Q(x;f,x_0) = f(x_0) + \langle \nabla f(x_0),x - x_0\rangle.
\end{equation}
This formulation provides an interpretation of ~\cite{simonyan14deep}'s saliency maps, which  visualize the gradient $S_1(x_0) = \nabla f(x_0)$ as an indication of salient image regions. They argue that large values of the gradient identify pixels that strongly affect the network output. However, an issue is that this interpretation \emph{breaks for a linear classifier}: If $f(x) = \langle w, x \rangle + b$, $S_1(x_0)=\nabla f(x_0) = w$ is independent of the image $x_0$ and hence cannot be interpreted as saliency.

The reason for this failure is that~\cref{e:simo} studies the variation of $f$ for arbitrary displacements $\Delta_x = x-x_0$ from $x_0$ and, for a linear classifier, the change is the same regardless of the starting point $x_0$. For a non-linear black box $f$ such as a neural network, this problem is reduced but not eliminated, and can explain why the saliency map $S_1$ is rather diffuse, with strong responses even where no obvious information can be found in the image (\cref{f:simonyan}).

We argue that the meaning of explanations depends in large part on the~\emph{meaning of varying the input $x$ to the black box}. For example, explanations in~\cref{s:meta} are based on letting $x$ vary in image category or in rotation. For saliency, one is interested in finding image regions that impact $f$'s output. Thus, it is natural to consider perturbations $x$ obtained by deleting subregions of $x_0$. If we model deletion by multiplying $x_0$ point-wise by a mask $m$, this amounts to studying the function $f(x_0 \odot m)$\footnote{$\odot$ is the Hadamard or element-wise product of vectors.}. The Taylor expansion of $f$ at $m=(1,1,\dots,1)$ is 
$
S_2(x_0) = \left.df(x_0 \odot m)/dm\right|_{m=(1,\dots,1)}= \nabla f(x_0) \odot x_0.
$
For a linear classifier $f$, this results in the saliency $S_2(x_0) = w \odot x_0$, which is large for pixels for which $x_0$ and $w$ are large simultaneously. We refine this idea for non-linear classifiers in the next section.
\section{Saliency revisited}\label{s:sal}

\begin{figure}\centering
\includegraphics[width=.7\linewidth]{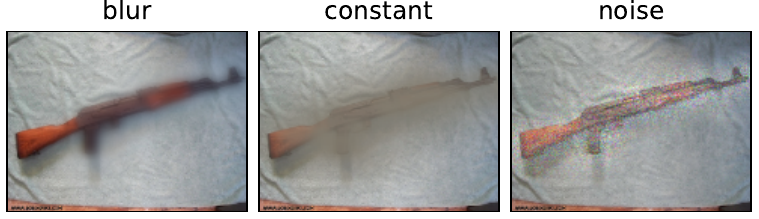}
\includegraphics[width=.7\linewidth]{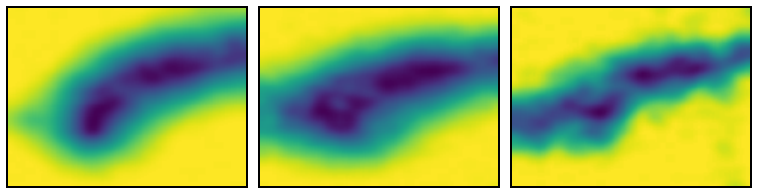}
\caption{Perturbation types. Bottom: perturbation mask; top: effect of blur, constant, and noise perturbations.}\label{f:perturb}
\end{figure}


\subsection{Meaningful image perturbations}\label{s:perturb}

In order to define an explanatory rule for a black box $f(x)$, one must start by specifying which variations of the input $x$ will be used to study $f$. The aim of saliency is to identify which regions of an image $x_0$ are used by the black box to produce the output value $f(x_0)$. We can do so by observing how the value of $f(x)$ changes as $x$ is obtained  ``deleting'' different regions $R$ of $x_0$. For example, if $f(x_0)=+1$ denotes a \emph{robin} image, we expect that $f(x)=+1$ as well unless the choice of $R$ deletes the robin from the image. Given that $x$ is a perturbation of $x_0$, this is a local explanation (\cref{s:local}) and we expect the explanation to characterize the relationship between $f$ and $x_0$.

While conceptually simple, there are several problems with this idea. The first one is to specify what it means ``delete'' information. As discussed in detail in~\cref{s:artifacts}, we are generally interested in simulating naturalistic or plausible imaging effect, leading to more meaningful perturbations and hence explanations. Since we do not have access to the image generation process, we consider three obvious proxies: replacing the region $R$ with a constant value, injecting noise, and blurring the image (\cref{f:perturb}).

Formally, let $m :\Lambda \rightarrow [0,1]$ be a \emph{mask}, associating each pixel $u \in \Lambda$ with a scalar value $m(u)$. Then the perturbation operator is defined as
$$
[\Phi(x_0;m)](u) =
 \begin{cases}
    m(u)x_0(u) + (1-m(u)) \mu_0, & \text{constant}, \\
    m(u)x_0(u) + (1-m(u)) \eta(u), & \text{noise}, \\
    \int g_{\sigma_0 m(u)}(v-u) x_0(v)\,dv, & \text{blur}, \\
 \end{cases}
$$
where $\mu_0$ is an average color, $\eta(u)$ are i.i.d.\ Gaussian noise samples for each pixel and $\sigma_0$ is the maximum isotropic standard deviation of the Gaussian blur kernel $g_\sigma$ (we use $\sigma_0 = 10$, which yields a significantly blurred image).

\subsection{Deletion and preservation}\label{s:minimal}

Given an image $x_0$, our goal is to summarize compactly the effect of deleting image regions in order to explain the behavior of the black box. One approach to this problem is to find deletion regions that are maximally informative.

In order to simplify the discussion, in the rest of the paper we consider black boxes $f(x)\in\mathbb{R}^C$ that generate a vector of scores for different hypotheses about the content of the image (e.g.\ as a softmax probability layer in a neural network). Then, we consider a ``deletion game'' where the goal is to find the smallest deletion mask $m$ that causes the score $f_c(\Phi(x_0;m)) \ll f_c(x_0)$ to drop significantly, where $c$ is the target class. Finding $m$ can be formulated as the following learning problem:
\begin{equation}\label{e:opt1}
m^\ast = \operatornamewithlimits{argmin}_{m \in [0,1]^\Lambda} \lambda \|\mathbf{1}-m\|_1 + f_c(\Phi(x_0;m))
\end{equation}
where $\lambda$ encourages most of the mask to be turned off (hence deleting a small subset of $x_0$). In this manner, we can find a highly informative region for the network.

One can also play an symmetric ``preservation game'', where the goal is to find the smallest subset of the image that must be retained to preserve the score $f_c(\Phi(x_0;m)) \geq f_c(x_0)$: 
$
m^\ast = \operatornamewithlimits{argmin}_m \lambda \|m\|_1 - f_c(\Phi(x_0;m))
$. 
The main difference is that the deletion game removes enough evidence to prevent the network from recognizing the object in the image, whereas the preservation game finds a minimal subset of sufficient evidence.

\paragraph{Iterated gradients.} Both optimization problems are solved by using a local search by means of gradient descent methods. In this manner, our method extracts information from the black box $f$ by computing its gradient, similar to the approach of~\cite{simonyan14deep}. However, it differs in that it extracts this information progressively, over several gradient evaluations, accumulating increasingly more information over time.

\subsection{Dealing with artifacts}\label{s:artifacts}

\begin{figure}\centering
  \includegraphics[width=0.9\linewidth,trim=0 .9em 0 .9em]{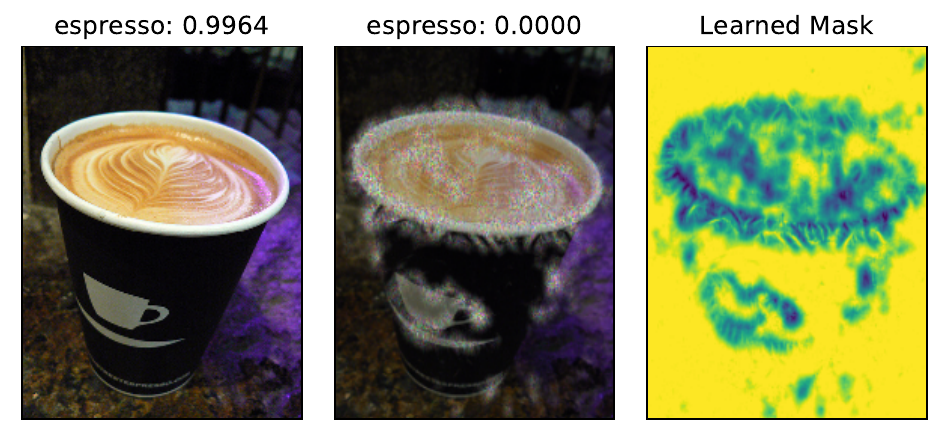}
  \includegraphics[width=0.9\linewidth,trim=0 .9em 0 .9em]{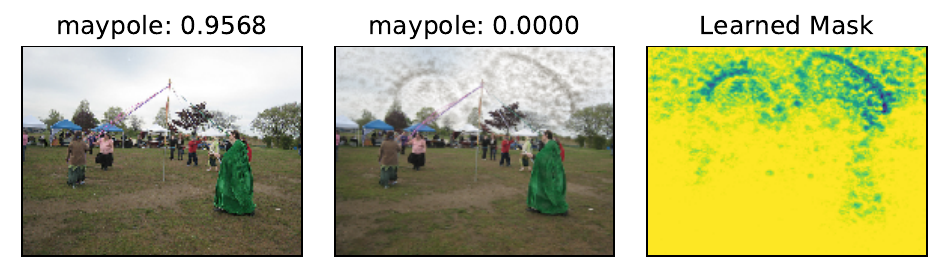}
\caption{From left to right: an image correctly classified with large confidence by GoogLeNet~\cite{szegedy2015going}; a perturbed image that is not recognized correctly anymore; the deletion mask learned with artifacts. Top: A mask learned by minimizing the top five predicted classes by jointly applying the constant, random noise, and blur perturbations. Note that the mask learns to add highly structured swirls along the rim of the cup ($\gamma = 1, {\lambda}_1 = 10^{-5}, {\lambda}_2 = 10^{-3}, \beta = 3$).
Bottom: A minimizing-top5 mask learned by applying a constant perturbation. Notice that the mask learns to introduce sharp, unnatural artifacts in the sky instead of deleting the pole ($\gamma = 0.1, {\lambda}_1 = 10^{-4}, {\lambda}_2 = 10^{-2}, \beta = 3$).}
\label{f:artifacts}
\end{figure}

By committing to finding a single representative perturbation, our approach incurs the risk of triggering artifacts of the black box. Neural networks, in particular, are known to be affected by surprising artifacts~\cite{kurakin2016adversarial,nguyen2015deep,mahendran2015understanding}; these works demonstrate that it is possible to find particular inputs that can drive the neural network to generate nonsensical or unexpected outputs. This is not entirely surprising since neural networks are trained discriminatively on natural image statistics. While not all artifacts look ``unnatural'', nevertheless they form a subset of images that is sampled with negligible probability when the network is operated normally.

Although the existence and characterization of artifacts is an interesting problem \emph{per se}, we wish to characterize the behavior of black boxes under normal operating conditions. Unfortunately, as illustrated in~\cref{f:artifacts}, objectives such as~\cref{e:opt1} are strongly attracted by such artifacts, and naively learn subtly-structured deletion masks that trigger them. This is particularly true for the noise and constant perturbations as they can more easily than blur create artifacts using sharp color contrasts (\cref{f:artifacts}, bottom row).


We suggests two approaches to avoid such artifacts in generating explanations. The first one is that powerful explanations should, just like any predictor, generalize as much as possible. For the deletion game, this means not relying on the details of a singly-learned mask $m$. Hence, we reformulate the problem to apply the mask $m$ stochastically, up to small random jitter.

Second, we argue that masks co-adapted with network artifacts are  \emph{not representative of natural perturbations}. As noted before, the meaning of an explanation depends on the meaning of the changes applied to the input $x$; to obtain a mask more representative of natural perturbations we can encourage it to have a simple, regular structure which cannot be co-adapted to artifacts. We do so by regularizing $m$ in total-variation (TV) norm and upsampling it from a low resolution version.

With these two modifications,~\cref{e:opt1} becomes:
\begin{multline}\label{e:opt2}
\min_{m\in[0,1]^\Lambda}
\lambda_1 \|\mathbf{1}-m\|_1
+
\lambda_2 \sum_{u\in\Lambda} \|\nabla m(u)\|^\beta_\beta \\
+ \mathbb{E}_\tau [ f_c(\Phi(x_0(\cdot - \tau),m)) ],
\end{multline}
where $M(v) = \sum_{u} g_{\sigma_m}(v/s - u) m(u)$. is the upsampled mask and $g_{\sigma_m}$ is a 2D Gaussian kernel. \Cref{e:opt2} can be optimized using stochastic gradient descent.


\paragraph{Implementation details.} Unless otherwise specified, the visualizations shown were generated using Adam~\cite{kingma2014adam} to minimize GoogLeNet's~\cite{szegedy2015going} softmax probability of the target class by using the blur perturbation with the following parameters: learning rate $\gamma = 0.1, N = 300$ iterations, ${\lambda}_1 = 10^{-4}, {\lambda}_2 = 10^{-2}, \beta = 3,$ upsampling a mask ($28\times 28$ for GoogLeNet) by a factor of $\delta = 8$, blurring the upsampled mask with $g_{\sigma_{m} = 5}$, and jittering the mask by drawing an integer from the discrete uniform distribution on $[0,\tau)$ where $\tau = 4$. We initialize the mask as the smallest centered circular mask that suppresses the score of the original image by $99\%$ when compared to that of the fully perturbed image, i.e. a fully blurred image.



\section{Experiments}\label{s:experiments}

\subsection{Interpretability}\label{s:interpretability}

\begin{figure}\centering
  \includegraphics[width=0.9\linewidth, trim=0 1em 0 1em]{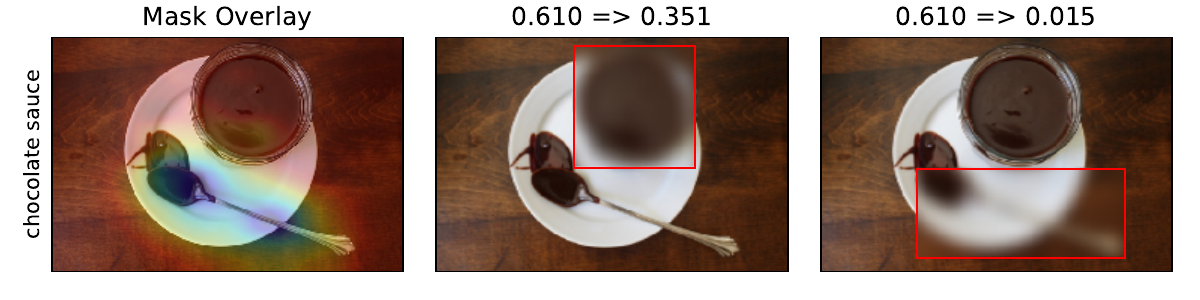}\\
  \includegraphics[width=0.9\linewidth, trim=0 1em 0 0]{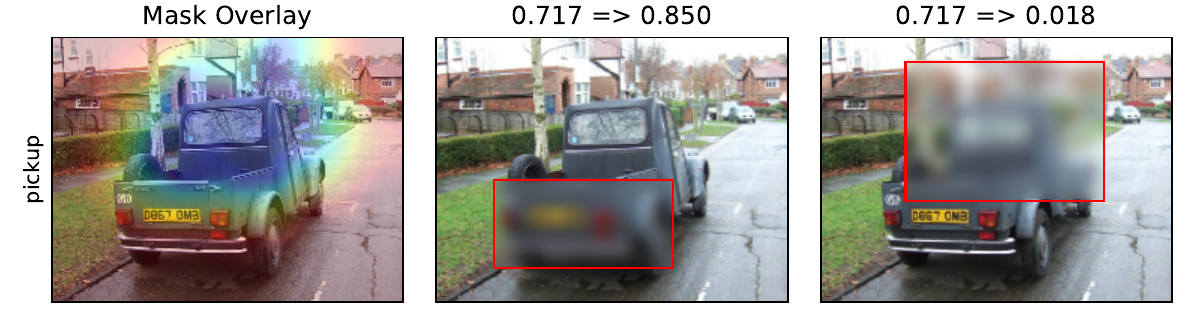}
   \caption{Interrogating suppressive effects. Left to right: original image with the learned mask overlaid; a boxed perturbation chosen out of interest (the truck's middle bounding box was chosen based on the contrastive excitation backprop heatmap from~\cref{f:comparison}, row 6); another boxed perturbation based on the learned mask (target softmax probabilities of  for the original and perturbed images are listed above).}
\label{f:manual_edit}
\end{figure}

An advantage of the proposed framework is that the generated visualizations are clearly interpretable. For example, the deletion game produces a minimal mask that prevents the network from recognizing the object.

When compared to other techniques (\cref{f:comparison}), this method can pinpoint the reason why a certain object is recognized without highlighting non-essential evidence. This can be noted in \cref{f:comparison} for the CD player (row 7) where other visualizations also emphasize the neighboring speakers, and similarly for the cliff (row 3), the street sign (row 4), and the sunglasses (row 8). Sometimes this shows that only a part of an object is essential: the face of the Pekenese dog (row 2), the upper half of the truck (row 6), and  the spoon on the chocolate sauce plate (row 1) are all found to be minimally sufficient parts.

While contrastive excitation backprop generated heatmaps that were most similar to our masks, our method introduces a quantitative criterion (i.e., maximally suppressing a target class score), and its verifiable nature (i.e., direct edits to an image), allows us to compare differing proposed saliency explanations and demonstrate that our learned masks are better on this metric. In~\cref{f:manual_edit}, row 2, we show that applying a bounded perturbation informed by our learned mask significantly suppresses the truck softmax score, whereas a boxed perturbation on the truck's back bumper, which is highlighted by contrastive excitation backprop in~\cref{f:comparison}, row 6, actually increases the score from $0.717$ to $0.850$.

The principled interpretability of our method also allows us to identify instances when an algorithm may have learned the wrong association. In the case of the chocolate sauce in~\cref{f:manual_edit}, row 1, it is surprising that the spoon is highlighted by our learned mask, as one might expect the sauce-filled jar to be more salient. However, manually perturbing the image reveals that indeed the spoon is more suppressive than the jar. One explanation is that the ImageNet ``chocolate sauce'' images contain more spoons than jars, which appears to be true upon examining some images. More generally, our method allows us to diagnose highly-predictive yet non-intuitive and possibly misleading correlations by identified machine learning algorithms in the data.


\subsection{Deletion region representativeness}\label{s:generality}

\begin{figure}[t]\centering
\includegraphics[width=0.9\linewidth]{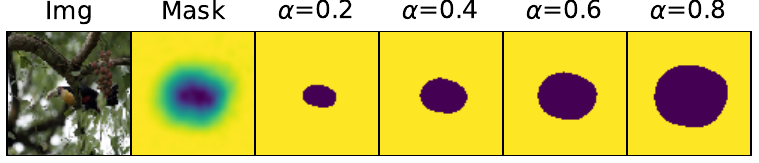}

\includegraphics[width=0.9\linewidth,trim=0 1em 0 0]{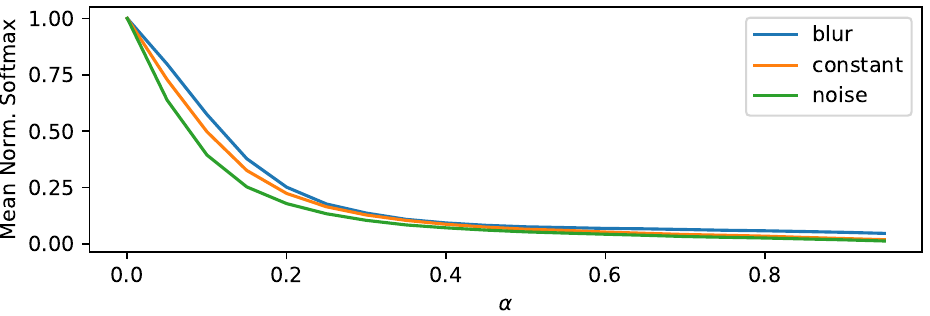}
   \caption{\textbf{(Top)} Left to right: original image, learned mask, and simplified masks for~\cref{s:generality} (not shown: further smoothed mask). \textbf{(Bottom)} Swift softmax score suppression is observed when using all three perturbations with simplified binary masks (top) derived from our learned masks, thereby showing the generality of our masks.}
\label{f:sanity_check_graph}
\end{figure}

To test that our learned masks are generalizable and robust against artifacts, we simplify our masks by further blurring them and then slicing them into binary masks by thresholding  the smoothed masks by $\alpha \in [0:0.05:0.95]$ (\cref{f:sanity_check_graph}, top; $\alpha \in [0.2,0.6]$ tends to cover the salient part identified by the learned mask). We then use these simplified masks to edit a set of 5,000 ImageNet images with constant, noise, and blur perturbations. Using GoogLeNet~\cite{szegedy2015going}, we compute normalized softmax probabilities\footnote{\label{n:normalize}$p'=\dfrac{p-p_0}{p_0-p_b}$, where $p,p_0,p_b$ are the masked, original, and fully blurred images' scores} (\cref{f:sanity_check_graph}, bottom). The fact that these simplified masks quickly suppress scores as $\alpha$ increases for all three perturbations gives confidence that the learned masks are identifying the right regions to perturb and are generalizable to a set of extracted masks and other perturbations that they were not trained on.

\subsection{Minimality of deletions}\label{s:deletion}

In this experiments we assess the ability of our method to correctly identify a minimal region that suppresses the object. Given the output saliency map, we normalize its intensities to lie in the range $[0,1]$, threshold it with $h \in [0:0.1:1]$, and fit the tightest bounding box around the resulting heatmap. We then blur the image in the box and compute the normalized\textsuperscript{\ref{n:normalize}} target softmax probability from GoogLeNet~\cite{szegedy2015going} of the partially blurred image.

\begin{figure}[t]\centering
\includegraphics[width=0.9\linewidth]{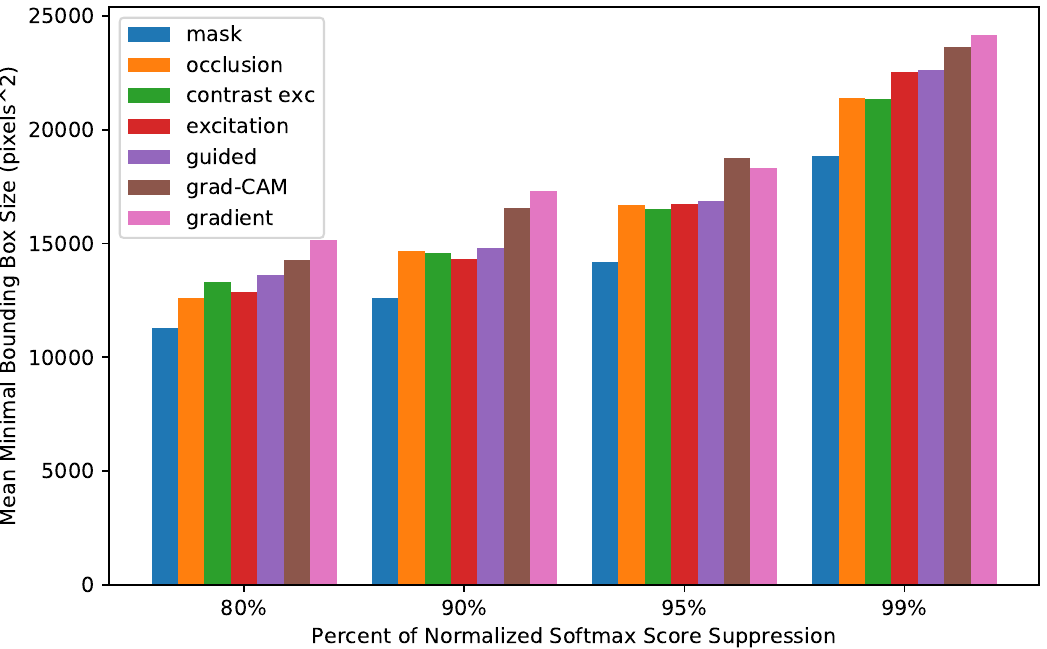}
   \caption{On average, our method generates the smallest bounding boxes that, when used to blur the original images, highly suppress their normalized softmax probabilities (standard error included).}
\label{f:deletion}
\end{figure}

From these bounding boxes and normalized scores, for a given amount of score suppression, we find the smallest bounding box that achieves that amount of suppression. \Cref{f:deletion} shows that, on average, our method yields the smallest minimal bounding boxes when considering suppressive effects of $80\%,90\%,95\%,\text{ and }99\%$. These results show that our method finds a small salient area that strongly impacts the network.

\subsection{Testing hypotheses: animal part saliency}\label{s:parts_exp}

\begin{figure}[t]\centering
  \includegraphics[width=0.9\linewidth]{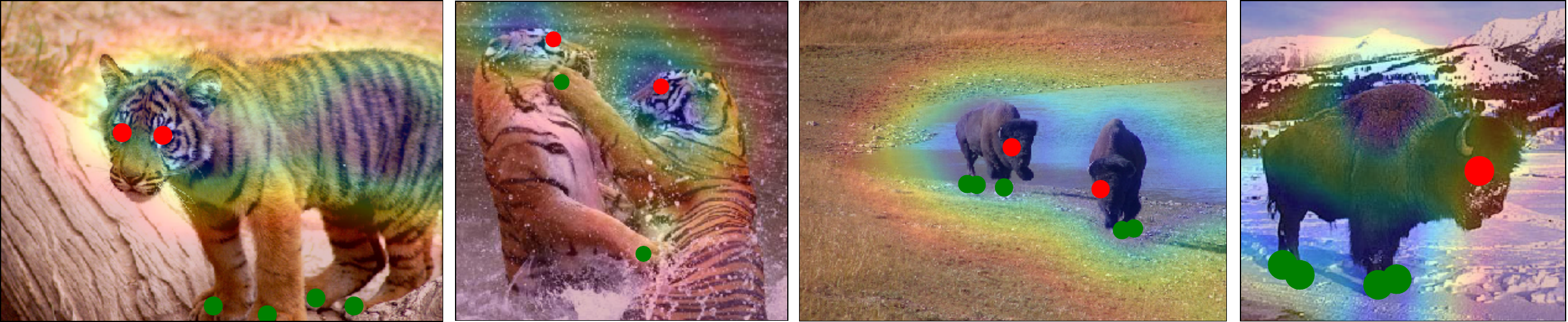}
   \caption{``tiger'' (left two) and ``bison'' (right two) images with eyes and feet annotations from~\cite{novotny2016have}; our learned masks are overlaid. The mean average feet:eyes intensity ratio for ``tigers'' ($N = 25$) is 3.82, while that for bisons ($N = 22$) is 1.07.}
\label{f:animal}
\end{figure}

From qualitatively examining learned masks for different animal images, we noticed that faces appeared to be more salient than appendages like feet. Because we produce dense heatmaps, we can test this hypothesis. From an annotated subset of the ImageNet dataset that identifies the keypoint locations of non-occluded eyes and feet of vertebrate animals~\cite{novotny2016have}, we select images from classes that have at least 10 images which each contain at least one eye and foot annotation, resulting in a set of 3558 images from 76 animal classes (\cref{f:animal}). For every keypoint, we calculate the average heatmap intensity of a $5\times 5$ window around the keypoint. For all 76 classes, the mean average intensity of eyes were lower and thus more salient than that of feet (see supplementary materials for class-specific results).

\subsection{Adversarial defense}
\begin{figure}[t]\centering
  \includegraphics[width=0.23\linewidth]{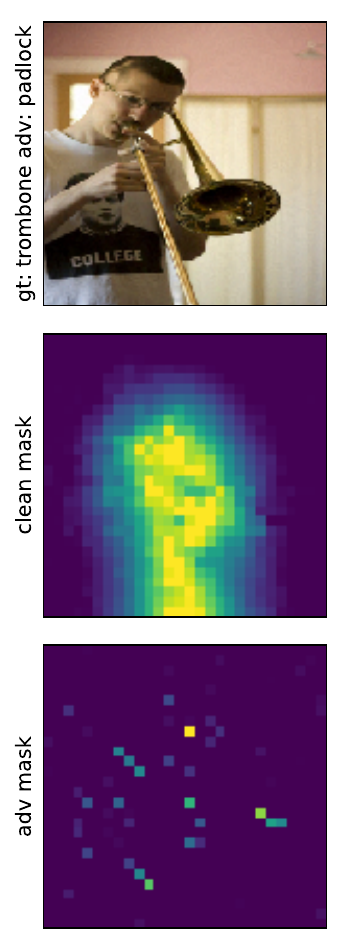}
  \includegraphics[width=0.67\linewidth]{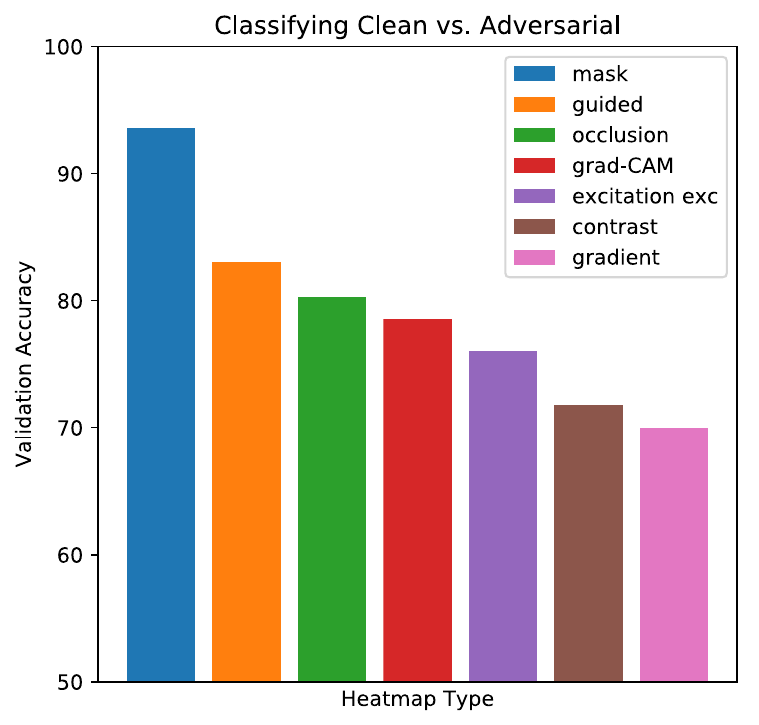}
   \caption{\textbf{(Left)} Difference between learned masks for clean (middle) and adversarial (bottom) images ($28 \times 28$ masks shown without bilinear upsampling). \textbf{(Right)} Classification accuracy for discriminating between clean vs. adversarial images using heatmap visualizations ($N_{trn} = 4000, N_{val} = 1000$).}
\label{f:adv}
\end{figure}

Adversarial examples~\cite{kurakin2016adversarial} are often generated using a complementary optimization procedure to our method that learns a imperceptible pattern of noise which causes an image to be misclassified when added to it. Using our re-implementation of the highly effective one-step iterative method ($\epsilon=8$)~\cite{kurakin2016adversarial} to generate adversarial examples, our method yielded visually distinct, abnormal masks compared to those produced on natural images (\cref{f:adv}, left). We train an Alexnet~\cite{krizhevsky2012imagenet} classifier (learning rate $\lambda_{lr} = 10^{-2}$, weight decay $\lambda_{L1} = 10^{-4}$, and momentum $\gamma = 0.9$) to distinguish between clean and adversarial images by using a given heatmap visualization with respect to the top predicted class on the clean and adversarial images (\cref{f:adv}, right); our method greatly outperforms the other methods and achieves a discriminating accuracy of $93.6\%$. 

Lastly, when our learned masks are applied back to their corresponding adversarial images, they not only minimize the adversarial label but often allow the original, predicted label from the clean image to rise back as the top predicted class. Our method recovers the original label predicted on the clean image 40.64\% of time and the ground truth label 37.32\% ($N = 5000$). Moreover, 100\% of the time the original, predicted label was recovered as one of top-5 predicted labels in the ``mask+adversarial'' setting. To our knowledge, this is the first work that is able to recover originally predicted labels without any modification to the training set-up and/or network architecture.

\subsection{Localization and pointing}

Saliency methods are often assessed in terms of weakly-supervised localization and a pointing game~\cite{zhang2016top}, which tests how discriminative a heatmap method is by calculating the precision with which a heatmap's maximum point lies on an instance of a given object class, for more harder datasets like COCO~\cite{lin2014microsoft}. Because the deletion game is meant to discover minimal salient part and/or spurious correlation, we do not expect it to be particularly competitive on localization and pointing but tested them for completeness.

For localization, similar to~\cite{zhang2016top,cao2015look}, we predict a bounding box for the most dominant object in each of $\sim$50k ImageNet~\cite{russakovsky2015imagenet} validation images and employ three simple thresholding methods for fitting bounding boxes. First, for value thresholding, we normalize heatmaps to be in the range of $[0,1]$ and then threshold them by their value with $\alpha \in [0:0.05:0.95]$. Second, for energy thresholding~\cite{cao2015look}, we threshold heatmaps by the percentage of energy their most salient subset covered with $\alpha \in [0:0.05:0.95]$. Finally, with mean thresholding~\cite{zhang2016top}, we threshold a heatmap by $\tau = \alpha \mu_{I}$, where $\mu_{I}$ is the mean intensity of the heatmap and $\alpha \in [0:0.5:10]$. For each thresholding method, we search for the optimal $\alpha$ value on a heldout set. Localization error was calculated as the IOU with a threshold of $0.5$.

~\Cref{t:localization} confirms that our method performs reasonably and shows that the three thresholding techniques affect each method differently. Non-contrastive, excitation backprop~\cite{zhang2016top} performs best when using energy and mean thresholding; however, our method performs best with value thresholding and is competitive when using the other methods: It beats gradient~\cite{simonyan14deep} and guided backprop~\cite{springenberg2014striving} when using energy thresholding; beats LRP~\cite{bach2015pixel}, CAM~\cite{zhou2016learning}, and contrastive excitation backprop~\cite{zhang2016top} when using mean thresholding (recall from~\cref{f:comparison} that the contrastive method is visually most similar to mask); and out-performs Grad-CAM~\cite{selvaraju2016grad} and occlusion~\cite{zeiler2014visualizing} for all thresholding methods.

\setlength{\tabcolsep}{3pt}
\begin{table}
\begin{center}
\begin{tabular}{|r||c|c||c|c||c|c|}
\hline
& 
\footnotesize{Val-$\alpha$*} &
\footnotesize{Err (\%)} &
\footnotesize{Ene-$\alpha$*} &
\footnotesize{Err} &
\footnotesize{Mea-$\alpha$*} &
\footnotesize{Err} \\
\footnotesize{Grad~\cite{simonyan14deep}} & \footnotesize{0.25} & \footnotesize{46.0} & \footnotesize{0.10} & \footnotesize{43.9} &  \footnotesize{5.0} &   \footnotesize{41.7${}^\mathsection$} \\
\footnotesize{Guid~\cite{springenberg2014striving,mahendran2016salient}} & \footnotesize{0.05} & \footnotesize{50.2} & \footnotesize{0.30} & \footnotesize{47.0} & \footnotesize{4.5} & \footnotesize{42.0${}^\mathsection$} \\
\footnotesize{Exc~\cite{zhang2016top}} & \footnotesize{0.15} & \footnotesize{46.1} & \footnotesize{0.60} & \footnotesize{\textbf{38.7}} & \footnotesize{1.5}  & \footnotesize{\textbf{39.0}${}^\mathsection$ }\\
\footnotesize{C Exc~\cite{zhang2016top}} & \footnotesize{---} & \footnotesize{---} & \footnotesize{---} & \footnotesize{---} & \footnotesize{0.0} & \footnotesize{57.0${}^\dagger$} \\
\footnotesize{Feed~\cite{cao2015look}} & \footnotesize{---} & \footnotesize{---} & \footnotesize{0.95} & \footnotesize{38.8${}^\dagger$} & \footnotesize{---} & \footnotesize{---} \\
\footnotesize{LRP~\cite{bach2015pixel}} & \footnotesize{---} & \footnotesize{---} & \footnotesize{---} & \footnotesize{---} & \footnotesize{1.0} & \footnotesize{57.8${}^\dagger$} \\
\footnotesize{CAM~\cite{zhou2016learning}}& \footnotesize{---} & \footnotesize{---} & \footnotesize{---} & \footnotesize{---} &\footnotesize{1.0} &\footnotesize{48.1${}^\dagger$}\\
\footnotesize{Grad-CAM~\cite{selvaraju2016grad}} & \footnotesize{0.30} & \footnotesize{48.1} & \footnotesize{0.70} & \footnotesize{48.0} &  \footnotesize{1.0} &   \footnotesize{47.5} \\
\footnotesize{Occlusion~\cite{zeiler2014visualizing}} & \footnotesize{0.30} & \footnotesize{51.2} & \footnotesize{0.55} & \footnotesize{49.4} &  \footnotesize{1.0} &   \footnotesize{48.6} \\
\footnotesize{Mask${}^\ddagger$} & \footnotesize{0.10}&\footnotesize{\textbf{44.0}}& \footnotesize{0.95}& \footnotesize{43.1}& \footnotesize{0.5}& \footnotesize{43.2} \\
\hline
\end{tabular}
\end{center}
\caption{Optimal $\alpha$ thresholds and error rates from the weak localization task on the ImageNet validation set using saliency heatmaps to generate bounding boxes. ${}^\dagger$Feedback error rate are taken from~\cite{cao2015look}; all others (contrastive excitation BP, LRP, and CAM) are taken from~\cite{zhang2016top}. ${}^\mathsection$Using~\cite{zhang2016top}'s code, we recalculated these errors, which are $\leq 0.4\%$ of the originally reported rates. ${}^\ddagger$Minimized top5 predicted classes' softmax scores and used $\lambda_1 = 10^{-3}$ and $\beta = 2.0$ (examples in supplementary materials).}
\label{t:localization}
\end{table}

For pointing,~\cref{t:pointing} shows that our method outperforms the center baseline, gradient, and guided backprop methods and beats Grad-CAM on the set of difficult images (images for which 1) the total area of the target category is less than $25\%$ of the image and 2) there are at least two different object classes). We noticed qualitatively that our method did not produce salient heatmaps when objects were very small. This is due to L1 and TV regularization, which yield well-formed masks for easily visible objects. We test two variants of occlusion~\cite{zeiler2014visualizing}, blur and variable occlusion, to interrogate if 1) the blur perturbation with smoothed masks is most effective, and 2) using the smallest, highly suppressive mask is sufficient (Occ${}^\mathsection$ and V-Occ in~\cref{t:pointing} respectively). Blur occlusion outperforms all methods except contrast excitation backprop while variable while variable occlusion outperforms all except contrast excitation backprop and the other occlusion methods, suggesting that our perturbation choice of blur and principle of identifying the smallest, highly suppressive mask is sound even if our implementation struggles on this task (see supplementary materials for examples and implementation details).

\setlength{\tabcolsep}{1pt}
\begin{table}
\begin{center}
\begin{tabular}{|r|c|c|c|c|c|c|c|c|c|c|}
\hline
& \footnotesize{Ctr} & \footnotesize{Grad}& \footnotesize{Guid} & \footnotesize{Exc} & \footnotesize{CExc} & \footnotesize{G-CAM} & \footnotesize{Occ} & \footnotesize{Occ${}^\mathsection$} &
\footnotesize{V-Occ${}^\dagger$} & \footnotesize{Mask${}^\ddagger$} \\
\hline
\footnotesize{All}
& \footnotesize{27.93} 
& \footnotesize{36.40}
& \footnotesize{32.68}
& \footnotesize{41.78}
& \footnotesize{\textbf{50.95}}
& \footnotesize{41.10}
& \footnotesize{44.50}
& \footnotesize{45.41}
& \footnotesize{42.31}
& \footnotesize{37.49}\\
\footnotesize{Diff}
& \footnotesize{17.86} 
& \footnotesize{28.21}
& \footnotesize{26.16}
& \footnotesize{32.73}
& \footnotesize{\textbf{41.99}}
& \footnotesize{30.59}
& \footnotesize{36.45}
& \footnotesize{37.45}
& \footnotesize{33.87}
& \footnotesize{30.64}\\
\hline
\end{tabular}
\end{center}
\caption{Pointing Game~\cite{zhang2016top} Precision on COCO Val Subset ($N \approx 20\text{k}$). ${}^\mathsection$Occluded with circles ($r = 35/2$) softened by $g_{\sigma_m = 10}$ and used to perturb with blur ($\sigma = 10$). ${}^\dagger$Occluded with variable-sized blur circles; from the top $10\%$ most suppressive occlusions, the one with the smallest radius is chosen and its center is used as the point. ${}^\ddagger$Used min. top-5 hyper-parameters ($\lambda_1 = 10^{-3}$, $\beta = 2.0$).}
\label{t:pointing}

\end{table}

\section{Conclusions}\label{s:conclusions}
We propose a comprehensive, formal framework for learning explanations as meta-predictors. We also present a novel image saliency paradigm that learns \emph{where} an algorithm \emph{looks} by discovering which parts of an image most affect its output score when perturbed. Unlike many saliency techniques, our method explicitly edits to the image, making it interpretable and testable. We demonstrate numerous applications of our method, 
%
%
and contribute new insights into the fragility of neural networks and their susceptibility to artifacts.
%

\paragraph{Acknowledgments.} We are grateful for the support by ERC StG 638009-IDIU.

{\small\bibliographystyle{ieee}\bibliography{refs}}
\end{document}